\documentclass[journal]{IEEEtran} 
\usepackage{amsmath,amsfonts}
\usepackage{algorithm}
\usepackage{algpseudocode}
\usepackage{array}
\usepackage[caption=false,font=normalsize,labelfont=sf,textfont=sf]{subfig}
\usepackage{textcomp}
\usepackage{stfloats}
\usepackage{url}
\usepackage{verbatim}
\usepackage{graphicx}
\usepackage{cite}
\usepackage{amsmath}
\usepackage{amssymb}
\usepackage{xcolor}
\usepackage{booktabs}
\usepackage{multirow}
\usepackage{orcidlink}
\usepackage[capitalise]{cleveref}
\usepackage{graphicx} 
\usepackage{caption} 
\usepackage{siunitx}
\begin{document}

\title{ReStNet: A Reusable \& Stitchable Network for Dynamic Adaptation on IoT Devices}

\author{Maoyu Wang\orcidlink{0009-0008-0633-6716}, Yao Lu\orcidlink{0000-0003-0655-7814}, ~\IEEEmembership{Student Member, IEEE}, Jiaqi Nie\orcidlink{0009-0003-7688-4391}, Zeyu Wang\orcidlink{0000-0001-7863-1070}, Yun Lin\orcidlink{0000-0003-1379-9301},~\IEEEmembership{Senior Member, IEEE}, Qi Xuan\orcidlink{0000-0002-6320-7012},~\IEEEmembership{Senior Member,~IEEE}, Guan Gui\orcidlink{0000-0003-3888-2881},~\IEEEmembership{Fellow,~IEEE}  
\thanks{This work was partially supported by the Key R\&D Program of Zhejiang under Grant 2022C01018 and by the National Natural Science Foundation of China under Grant U21B2001, 62301492 and 61973273. (Corresponding author: Qi Xuan)}
\thanks{Maoyu Wang is with the College of Computer science and Technology, Zhejiang University of Technology, Hangzhou, China (e-mail: wangmy.zjut@gmail.com)}
\thanks{Yao Lu is with the Institute of Cyberspace Security, College of Information Engineering, Zhejiang University of Technology, Hangzhou 310023, China, with the Binjiang Institute of Artificial Intelligence, Zhejiang University of Technology, Hangzhou 310056, China, also with the Centre for Frontier AI Research, Agency for Science, Technology and Research, Singapore 138632 (e-mail: yaolu.zjut@gmail.com).}
\thanks{Jiaqi Nie, Zeyu Wang and Qi Xuan are with the Institute of Cyberspace Security, College of Information Engineering, Zhejiang University of Technology, Hangzhou 310023, China, also with the Binjiang Institute of Artificial Intelligence, Zhejiang University of Technology, Hangzhou 310056, China (e-mail: jiaqinie@zjut.edu.cn, vencent\_wang@outlook.com, xuanqi@zjut.edu.cn).}
\thanks{Yun Lin is with the College of Information and Communication Engineering, Harbin Engineering University, Harbin, China (e-mail: linyun@hrbeu.edu.cn).}
\thanks{Guan Gui is with the College of Telecommunications and Information Engineering, Nanjing University of Posts and Telecommunications, Nanjing 210003, China (e-mail: guiguan@njupt.edu.cn).}}


\markboth{Journal of \LaTeX\ Class Files,~Vol.~14, No.~8, August~2021}%
{Shell \MakeLowercase{\textit{et al.}}: A Sample Article Using IEEEtran.cls for IEEE Journals}


\maketitle
\begin{abstract}

With the rapid development of deep learning, a growing number of pre-trained models have been publicly available. However, deploying these fixed models in real-world IoT applications is challenging because different devices possess heterogeneous computational and memory resources, making it impossible to deploy a single model across all platforms. Although traditional compression methods---such as pruning, quantization, and knowledge distillation---can improve efficiency, they become inflexible once applied and cannot adapt to changing resource constraints. To address these issues, we propose ReStNet, a \textbf{Re}usable and \textbf{St}itchable \textbf{Net}work that dynamically constructs a hybrid network by stitching two pre-trained models together. Implementing ReStNet requires addressing several key challenges, including how to select the optimal stitching points, determine the stitching order of the two pre-trained models, and choose an effective fine-tuning strategy. To systematically address these challenges and adapt to varying resource constraints, ReStNet determines the stitching point by calculating layer-wise similarity via Centered Kernel Alignment (CKA). It then constructs the hybrid model by retaining early layers from a larger-capacity model and appending deeper layers from a smaller one. To facilitate efficient deployment, only the stitching layer is fine-tuned. This design enables rapid adaptation to changing budgets while fully leveraging available resources. Moreover, ReStNet supports both homogeneous (CNN-CNN, Transformer-Transformer) and heterogeneous (CNN-Transformer) stitching, allowing to combine different model families flexibly. Extensive experiments on multiple benchmarks demonstrate that ReStNet achieve flexible accuracy–efficiency trade‐offs at runtime while significantly reducing training cost.

\end{abstract}


\section{Introduction}

\IEEEPARstart{W}{ith} the rapid development of deep learning, the community has witnessed an explosion of publicly available pre-trained models hosted on platforms such as Hugging Face~\cite{huggingface} and timm~\cite{timm}. These models cover a wide range of sizes and structures, from lightweight models for mobile devices to large models designed for high-performance servers~\cite{ResNet,VGG,swin,deit,vit,ding2022davit,ryali2023hiera}. 
Although these pre-trained models have greatly reduced the training cost and development effort, deploying them in real-world IoT scenarios introduces new challenges. IoT devices differ significantly in computational capability and memory availability, making it impossible for a single fixed model to fit all platforms.

In recent years, many researchers have turned to model compression techniques as a way to tailor powerful models to the limited resources of edge devices. The most widely adopted model compression techniques include pruning~\cite{pruning,lu2022understanding,lu2024generic,lu2024redtest,li2024sglp,lu2025fcos,li2025sepprune,lu2024reassessing}, knowledge distillation~\cite{distillation,hinton2015distilling,distillation2,distillation3,distillation4,distillation5}, and quantization\cite{quantization,quantization2,quantization3,quantization4,quantization5,yang2019quantization}. These methods effectively reduce the model size and computational overhead while maintaining competitive performance. However, they all rely on a fixed architecture: once compression is applied, the structural changes are irreversible. Therefore, these techniques lack the flexibility to customize a single compression model to be deployed in IoT devices with different computing power and memory capacity, thus limiting their applicability in heterogeneous IoT environments. This limitation raises an important question: can we design a method that adapts to the heterogeneous resource constraints of various IoT devices instead of relying on a fixed model?

To this end, we design ReStNet (a \textbf{Re}usable and \textbf{St}itchable \textbf{Net}work) that combines components from different pre-trained models into a single hybrid network via model stitching~\cite{modelStitching}, enabling flexible adaptation to diverse resource constraints in IoT devices. While promising, applying model stitching in practice introduces three core problems:
\begin{itemize}
\item[1.] \textbf{Stitching‐point Selection:} Identifying an appropriate stitching point is non-trivial, as different combinations can lead to drastically different performance. A brute‐force search over all possible layer pairs between two models is prohibitive. For example, stitching two $100$-layer models would require evaluating nearly $10{,}000$ potential connections.
\item[2.] \textbf{Stitching‐order Determination:} After determining the best stitching point, how to seamlessly combine the partial models of two independently pre-trained models remains a tricky problem. Should the larger model components be stitched before or after the smaller model components?
\item[3.] \textbf{Fine‐tuning Strategy Choice:} Once a stitching point and order are fixed, one must decide which weights to update, whether to fine‐tune only the stitching layer or to fine‐tune the entire model. This decision directly impacts the deployment overhead.
\end{itemize}

To address these challenges, ReStNet adopts the following strategy. Under a given resource budget, ReStNet first computes Centered Kernel Alignment (CKA) similarity scores for every layer pair between two pre-trained models and selects the pair with the highest score as the optimal stitching point. By selecting the most semantically consistent layers, we avoid an exhaustive search over all combinations, thus significantly reducing the search cost. Then ReStNet constructs the hybrid network by retaining the early layers of the model with greater capacity up to that point, then appending the later layers of the smaller model. Finally, to minimize retraining overhead and enable fast adaptation, only a single concatenated layer is fine-tuned after assembling the hybrid network, while all other weights remain frozen. In this way, ReStNet provides solid technical support for the efficient deployment of deep learning models across heterogeneous IoT devices with diverse computational capabilities.

To evaluate the effectiveness of ReStNet, we conduct a comprehensive evaluation on five widely used benchmark datasets: CIFAR10~\cite{CIFAR}, CIFAR100~\cite{CIFAR}, DTD~\cite{DTD}, Oxford-IIIT Pets~\cite{Pets}, and ImageNette~\cite{Howard_Imagenette_2019}. Our experiments cover both CNNs (ResNet~\cite{ResNet}, VGG~\cite{VGG}) and transformers (Swin~\cite{swin}, DeiT~\cite{deit}, and include two stitching types: homogeneous stitching (CNN-CNN and Transformer-Transformer) as well as heterogeneous stitching (CNN-Transformer). Across all datasets, ReStNet achieves a flexible trade-off between accuracy and efficiency at runtime with extremely low training cost.


To summarize, our main contributions are three-fold:
\begin{itemize}
\item[$\bullet$] \textbf{A New Model for Dynamic Resource Adaptation:} We introduce ReStNet, a reusable and stitchable model specifically designed to address the heterogeneous computational and memory constraints of IoT devices. By combining components from large‐capacity and small‐capacity pre-trained models, ReStNet enables a single hybrid model to flexibly span a wide range of resource budgets, eliminating the need for multiple fixed models.
\item[$\bullet$] \textbf{Support for Heterogeneous and Homogeneous Stitching:} ReStNet supports both homogeneous (e.g., CNN–CNN or Transformer–Transformer) and heterogeneous (e.g., CNN–Transformer) stitching. By enabling hybrid designs across different architectures, ReStNet unlocks novel combinations that leverage the strengths of multiple model families.
\item[$\bullet$] \textbf{Extensive Experimental Validation:} Extensive experiments across $5$ benchmarks demonstrate that, compared to training individual models from scratch, ReStNet can achieve flexible accuracy–efficiency trade‐offs at runtime while significantly reducing training cost.
\end{itemize}
In the remainder of this paper, we first introduce related works in \cref{sec:Related Works}. Preliminaries are presented in \cref{sec:Preliminaries}. Our method is detailed in \cref{sec:Method}, and all relevant experiments are discussed in \cref{sec:Experiments}. Finally, the paper concludes in \cref{sec:Conclusion}.

\section{Related Works}
\label{sec:Related Works}
\subsection{Neural Representation Similarity}
Exploring neural representations helps reveal how models process data, offering insights into their functional behavior. This has led to the development of neural representation similarity methods, which aim to expose the internal dynamics of models and compare their functional properties. A variety of techniques have been proposed to quantify the similarity between neural representations. Early efforts focus on linear regression~\cite{hinton2015distilling,Linear2}. Subsequently, Canonical Correlation Analysis (CCA)~\cite{CCA} and its variants, such as SVCCA~\cite{SVCCA}, introduce statistical alignment over projected subspaces. Later, Centered Kernel Alignment (CKA)~\cite{CKA} has been proposed, offering invariance to orthogonal transformations and demonstrating stronger alignment with human perceptions of functional similarity. More recently,~\cite{GBS} presents Graph-Based Similarity (GBS), measuring the similarity based on graph structures.

Nowadays, representation similarity has been widely applied in practical tasks such as model distillation~\cite{similarity_distillation}, pruning~\cite{lu2024redtest}, and transfer learning~\cite{similar_transfer,similar_transfer2}. In this work, we further explore its potential in the context of model stitching. Specifically, we adopt CKA to precisely quantify the functional similarity between various layers.

\subsection{Model Stitching}


Model stitching builds a hybrid network by stitching different parts of different pre-trained models, thereby reusing existing models. It was originally proposed to study the alignment of representations between models~\cite{modelStitching}. For example, early studies show that stitching two models with the same architecture but different initialization can achieve good performance~\cite{stitching_same}, while later studies extended this idea to heterogeneous architectures, such as combining CNN with Transformer~\cite{stitching_differ,resnet_vit}.

Nowadays, model stitching has developed into a practical tool for building composite models. However, most existing approaches rely on manual selection or grid search to determine stitching points~\cite{SNnet}, which limits their scalability and flexibility. In this paper, we introduce ReStNet, which leverages CKA-based similarity to automatically identify the best stitching points, thus extending the model stitching capabilities. ReStNet can stitch homogeneous and heterogeneous architectures to obtain diverse models that support flexible deployment under different resource budgets, which is particularly suitable for IoT applications.

\section{Method}
In this section, we first introduce the preliminary of model stitching at \cref{sec:Preliminaries}. Next, we elaborate on the details of our proposed ReStNet at \cref{sec:Method}.

\subsection{Preliminaries of Model Stitching}
\label{sec:Preliminaries}
Model stitching is a technique that constructs new model architectures by stitching together components of existing pre-trained models. It inherits the advantages of the original models and enables scalable model structure designs tailored to various hardware requirements. Specifically, given two pre-trained models $f^{(1)}\left(x ; \theta^{(1)}\right)$ and $f^{(2)}\left(x ; \theta^{(2)}\right)$:
\begin{equation}
\begin{aligned}
f^{(1)}\left(x ; \theta^{(1)}\right)=&f_k^{(1)}  \circ \cdots \circ f_{m+1}^{(1)} \circ f_{m}^{(1)} \circ \cdots \circ f_1^{(1)}(x), \\
f^{(2)}\left(x ; \theta^{(2)}\right)=&f_l^{(2)} \circ \cdots \circ f_{n+1}^{(2)} \circ f_{n}^{(2)} \circ \cdots \circ f_1^{(2)}(x), \\
& \text{s.t.,} \quad k > m > 1, \quad l > n > 1,
\end{aligned}
\end{equation}
where $f_i^{(j)}$ represents the function of the i-th layer of model $j$, $\circ$ indicates function composition, $x$ denotes the input data and $\theta^{(j)}$ is the parameters of model $j$. The basic idea of model stitching involves splitting $f^{(1)}\left(x ; \theta^{(1)}\right)$ and $f^{(2)}\left(x ; \theta^{(2)}\right)$ into two parts at layer indexes $m$ and $n$, respectively:
\begin{equation}
\begin{aligned}
& f^{(1)}\left(x ; \theta^{(1)}\right)=\left(f_k^{(1)}  \circ \cdots \circ f_{m+1}^{(1)} \right) \circ \left( f_{m}^{(1)} \circ \cdots \circ f_1^{(1)}\right)(x), \\
& f^{(2)}\left(x ; \theta^{(2)}\right)=\left(f_l^{(2)} \circ \cdots \circ f_{n+1}^{(2)} \right) \circ \left( f_{n}^{(2)} \cdots \circ f_1^{(2)}\right)(x).
\end{aligned}
\end{equation}
Next, suppose we want to combine the first half of pre-trained model $f^{(1)}\left(x ; \theta^{(1)}\right)$ (i.e., $f_{m}^{(1)} \circ \cdots \circ f_1^{(1)}$) with the second half of pre-trained model $f^{(2)}\left(x ; \theta^{(2)}\right)$ (i.e., $f_l^{(2)} \circ \cdots \circ f_{n+1}^{(2)}$) to create a new hybrid model $g(x)$. It is worth noting that due to the differences in model architecture, the feature dimensions at the merging point are usually inconsistent, making it infeasible to merge the two models directly. To address this issue, we introduce a stitching layer $\mathcal{S}$ to adjust the feature size:
\begin{equation}
g(x)=\left(f_{l}^{(2)} \circ \cdots \circ f_{n+1}^{(2)}\right) \circ \mathcal{S} \circ \left(f_{m}^{(1)} \circ \cdots \circ f_1^{(1)}\right)(x).
\label{eq3}
\end{equation}
For ease of writing, we rewrite \cref{eq3} as $g(x)=\left(f_{l:n+1}^{(2)} \circ \mathcal{S} \circ f_{m:1}^{(1)} \right)(x)$, where $f_{m:1}^{(1)}$ denotes the first $m$ layers of $f^{(1)}\left(x ; \theta^{(1)}\right)$, $f_{l:n+1}^{(2)}$ denotes the last $l-n$ layers of $f^{(2)}\left(x ; \theta^{(2)}\right)$. By controlling the splitting point $m$ and $n$, model stitching can produce a sequence of stitched models.

\begin{figure*}[t]
	\centering
    \centering
    \includegraphics[width=0.99\textwidth]{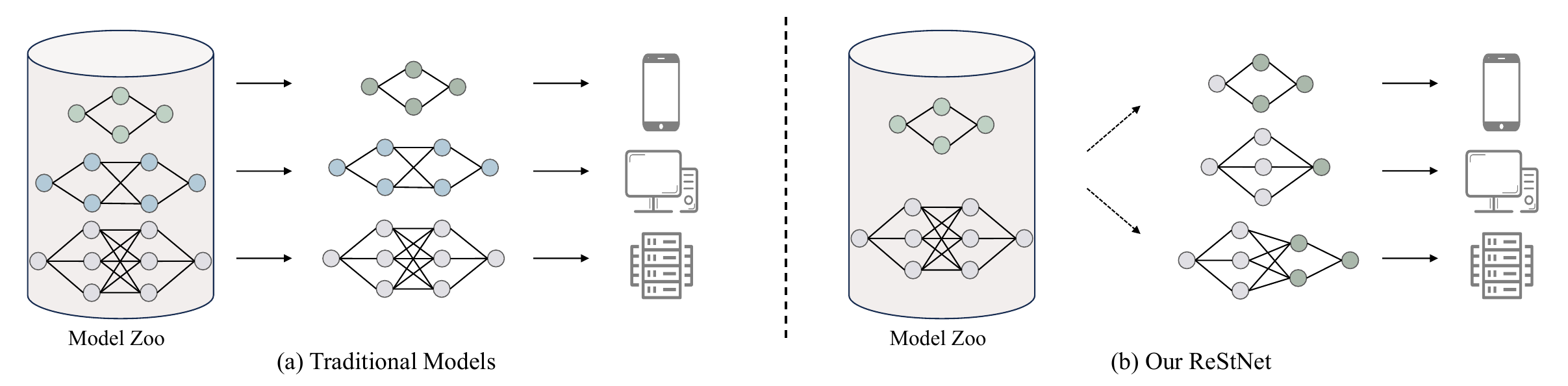} 
 \caption{Comparison of traditional fixed models and our ReStNet. (a) shows the traditional deployment method. A single fixed model only targets a specific edge device and cannot adapt to changing resource constraints. (b) ReStNet is introduced. By stitching pre-trained models, ReStNet can be reconfigured to adapt to different edge devices with dynamically changing computation, memory, and energy constraints.}
 \label{fig:framework}
\end{figure*}

\subsection{ReStNet}
\label{sec:Method}
Inspired by the insights gained from model stitching, we introduce ReStNet, a reusable and stitchable network, as illustrated in \cref{fig:framework}. ReStNet is inspired by the growing number of pre-trained models in the deep learning field, where most well-trained models cannot adapt to scenarios with dynamic resource constraints. To this end, ReStNet introduces a lightweight stitching layer to merge different pre-trained models to generate a series of stitched models that can adapt to different computing overhead requirements. In the following, we will elaborate on the implementation method in detail, including where and how to stitch, as well as an effective and efficient training strategy for ReStNet.

\begin{figure*}[t]
	\centering
    \centering
    \includegraphics[width=0.99\textwidth]{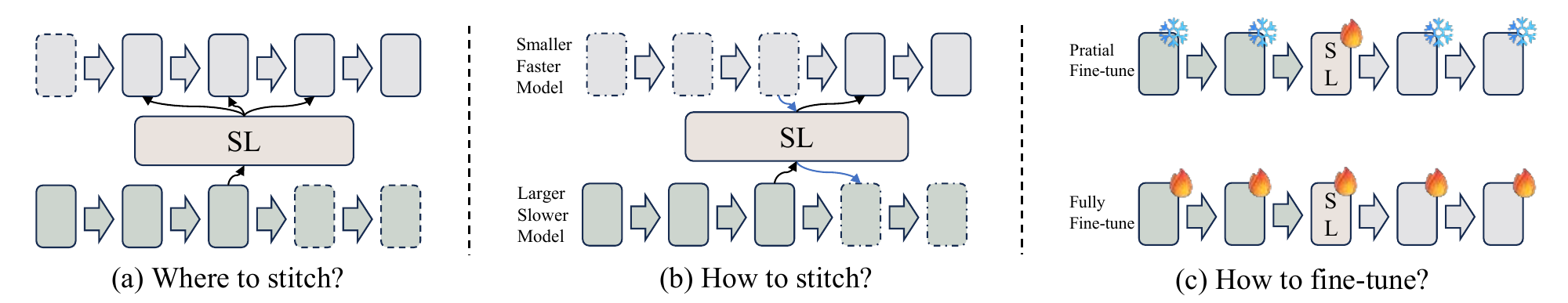} 
 \caption{Illustrations of the core challenges of ReStNet. (a) indicates that there are numerous choices for where to stitch between two pre-trained models. (b) introduces two stitching directions: Fast-to-Slow and Slow-to-Fast. (c) shows the partial and fully fine-tuning strategies. ``SL'' means the stitching layer.}
 \label{fig:vis}
\end{figure*}

\textbf{Where to stitch: the choice of layer indexes $m$ and $n$.} When splicing two different models, how to choose the optimal split point of two pre-trained models (i.e., the choice of layer indexes $m$ and $n$) is a key issue (as illustrated in \cref{fig:vis}(a)). The selection of split points not only needs to ensure feature compatibility and dimensional matching, but also needs to comprehensively consider the limitations of computing resources. Therefore, the above requirements can be formalized as the following formula:                    
\begin{equation}
\begin{aligned}
g^*(x)=\max P_T\left(f_{l:n+1}^{(2)} \circ \mathcal{S} \circ f_{m:1}^{(1)} \right), \\ 
\text{s.t.,} \quad \text{cost}\left(f_{l:n+1}^{(2)} \circ \mathcal{S} \circ f_{m:1}^{(1)}\right) \leq \epsilon,
\end{aligned}
\end{equation}
where $P_T()$ denotes the performance on a given task $T$, $\text{cost}\left(f_{l:n+1}^{(2)} \circ \mathcal{S} \circ f_{m:1}^{(1)}\right) \leq \epsilon$ ensures that the computational overhead of the stitched model must be under certain computational constraints $\epsilon$. Here, we take $f_{l:n+1}^{(2)} \circ \mathcal{S} \circ f_{m:1}^{(1)}$ as an example. We can also use $f_{k:m+1}^{(1)} \circ \mathcal{S} \circ f_{n:1}^{(2)}$ as the stitched model. In the subsection "How to stitch: the stitching directions", we will explain whether to choose $f_{l:n+1}^{(2)} \circ \mathcal{S} \circ f_{m:1}^{(1)}$ or $f_{k:m+1}^{(1)} \circ \mathcal{S} \circ f_{n:1}^{(2)}$.

Given that the feature representations and dimensions of two different models at different layers may differ significantly, in this paper, we select the split points of the pre-trained models based on representation similarity. Selecting layers with higher representation similarity as split points for model stitching can effectively reduce the information loss caused by feature mismatch, thereby improving the performance and training stability of the stitched model. Thanks to the effectiveness of Centered Kernel Alignment~\cite{gretton2005measuring} (CKA) in measuring representation similarity, we utilize it as our measurement metric.


Specifically, given a batch of examples $X_{b}$, the feature representations of $F_{m}^{(1)} \in \mathbb{R} ^{b\times c_m \times h_m \times w_m}$ and $F_{n}^{(2)} \in \mathbb{R} ^{b\times c_n \times h_n \times w_n}$ can be formulated as:
\begin{equation}
\begin{aligned}
F_{m}^{(1)} &= f_{m:1}^{(1)}(X_{b}), \\
F_{n}^{(2)} &= f_{n:1}^{(2)}(X_{b}),
\end{aligned}
\end{equation}
where $b$, $c$, $h$ and $w$ denote the number of samples, the number of channels, height and width, respectively. Then we apply a $\operatorname{flatten}$ operation to $F_{m}^{(1)}$ and $F_{n}^{(2)}$, preserving the number of samples $b$ and combining the dimensions $c$, $h$, and $w$ into a single dimension. Now $F_{m}^{(1)}$ and $F_{n}^{(2)}$ are transformed into $\mathbb{R} ^{b\times \left(c_m \times h_m \times w_m\right)}$ and $\mathbb{R} ^{b\times \left(c_n \times h_n \times w_n\right)}$. Based on the obtained feature representation, we calculate the gram matrices $K = F_{m}^{(1)} {F_{m}^{(1)}}^{\mathsf{T}}$ and $L = F_{n}^{(2)} {F_{n}^{(2)}}^{\mathsf{T}}$, which reflect the similarities between a pair of examples. Subsequently, we utilize Hilbert-Schmidt Independence Criterion~\cite{gretton2005measuring} ($\operatorname{HSIC}$) to calculate the statistical independence between $K$ and $L$:
\begin{equation}
\begin{aligned}
\operatorname{HSIC}(F_{m}^{(1)}, F_{n}^{(2)})&=\\
&\frac{1}{(b-1)^2} \operatorname{tr}\left(F_{m}^{(1)} {F_{m}^{(1)}}^{\mathsf{T}} \cdot H \cdot F_{n}^{(2)} {F_{n}^{(2)}}^{\mathsf{T}} \cdot H\right),
\end{aligned}
\end{equation}
where $H=I_b-\frac{1}{b} \mathbf{1 1}^T$ is the centering matrix. It is worth noting that $\operatorname{HSIC}$ is not invariant to isotropic scaling, but it can be made invariant through normalization. Therefore, CKA can be calculated as follows: 
\begin{equation}
\begin{aligned}
&\operatorname{CKA}(F_{m}^{(1)}, F_{n}^{(2)}) = \\
&\frac{\operatorname{HSIC}(F_{m}^{(1)} {F_{m}^{(1)}}^{\mathsf{T}}, F_{n}^{(2)} {F_{n}^{(2)}}^{\mathsf{T}})}{\sqrt{\operatorname{HSIC}(F_{m}^{(1)} {F_{m}^{(1)}}^{\mathsf{T}}, F_{m}^{(1)} {F_{m}^{(1)}}^{\mathsf{T}})\operatorname{HSIC}(F_{n}^{(2)} {F_{n}^{(2)}}^{\mathsf{T}}, F_{n}^{(2)} {F_{n}^{(2)}}^{\mathsf{T}})}}.
\label{eq:cka}
\end{aligned}
\end{equation}Subsequently, for each layer $i$ of $f^{(1)}\left(x ; \theta^{(1)}\right)$ and each layer $j$ of $f^{(2)}\left(x ; \theta^{(2)}\right)$, we calculate their CKA similarity and combine these values into a similarity matrix $\mathcal{S}$, where $ \mathcal{S}_{ij}=\operatorname{CKA}(F_{i}^{(1)}, F_{j}^{(2)})$. Then we select the layer pair $\{\left(i_{\text{top}}, j_{\text{top}}\right)\}$ with the highest similarity scores from the similarity matrix $\mathcal{S}$ and calculate the total number of parameters after stitching:
\begin{equation}
\begin{aligned}
    \text{Params}&\left(i_{\text{top}}, j_{\text{top}}\right)=\\
    &\text{Params}\left(f_{l:j_{\text{top}}+1}^{(2)}\right)+\text{Params}\left(\mathcal{S}\right)+\text{Params}\left(f_{i_{\text{top}}:1}^{(1)} \right),
\end{aligned}
\end{equation}
where $\text{Params}\left(f_{l:j_{\text{top}}+1}^{(2)}\right)$ denotes the number of parameters from layer $j_{\text{top}}+1$ to $l$ of model $f^{(2)}\left(x ; \theta^{(2)}\right)$, $\text{Params}\left(f_{i_{\text{top}}:1}^{(1)} \right)$ denotes the number of parameters from layer $1$ to $i_{\text{top}}$ of model $f^{(1)}\left(x ; \theta^{(1)}\right)$, $\text{Params}\left(\mathcal{S}\right)$ represents the number of parameters of stitching layer $\mathcal{S}$. If the candidate layer pair $\{\left(i_{\text{top}}, j_{\text{top}}\right)\}$ satisfies the resource constraint $\epsilon$ (i.e., $\text{Params}\left(i_{\text{top}}, j_{\text{top}}\right)\leq \epsilon$), we select it as the final splicing point $\{\left(i^*_{\text{top}}, j^*_{\text{top}}\right)\}$. Otherwise, we proceed with the second, third, and so on, most similar ones until the computational constraints are met. Finally, we obtained a stitching model under the resource constraint $\epsilon$:
\begin{equation}
    g(x)=f_{l:j^*_{\text{top}}+1}^{(2)} \circ \mathcal{S} \circ f_{i^*_{\text{top}}:1}^{(1)}.
\end{equation}The overall algorithm flow is shown in \cref{algorithm:Stitching}. It is worth noting that the number of parameters is widely used as a metric to evaluate model complexity and resource requirements, such as in model compression~\cite{shan2023low,yang2022mobilenet,lin2020improved,lin2023glr,lu2024generic,chen2023rgp} and lightweight model design~\cite{zhang2023lightweight,lu2024redtest,tu2020complex}. Therefore, in this paper, we choose the number of parameters as the default resource constraint metric, though we will demonstrate in \cref{sec:additional} that other choices are also possible.

\begin{algorithm}[t]

    \caption{Model Stitching}
    \label{algorithm:Stitching}
    \textbf{Input}: A batch of examples $X_{b}$, two pre-trained models: $f^{(1)}\left(x ; \theta^{(1)}\right)$ with $k$ layers and $f^{(2)}\left(x ; \theta^{(2)}\right)$ with $l$ layers, as well as the predefined computational overhead $\epsilon$ of the stitched model.\\
    \textbf{Output}: A stitched model $g(x)$.\\
    \begin{algorithmic}[1] 
    \State Initializing $\mathcal{S}=0_{k,l}$
    \For{$i = 1$ to $k$}
        \State $F_{i}^{(1)} = f_{i:1}^{(1)}(X_{b})$
        \State $F_{i}^{(1)} = \operatorname{flatten}(F_{i}^{(1)})$
        \For{$j = 1$ to $l$}
            \State $F_{j}^{(2)} = f_{j:1}^{(2)}(X_{b})$
            \State $F_{j}^{(2)} = \operatorname{flatten}(F_{j}^{(2)})$
            \State \text{Using \cref{eq:cka} to calculate the CKA value $u$.}
            \State $\mathcal{S}_{i.j}=u$
        \EndFor
    \EndFor
    \State Initialize a flag $\text{found} = \text{False}$
    \While{$\text{found} == \text{False}$}
    \State Find $\left(i_{\text{top}}, j_{\text{top}}\right) = \arg\max_{i,j} \mathcal{S}_{ij}$ 
    \State $\text{Params}\left(i_{\text{top}}, j_{\text{top}}\right)=\text{Params}\left(f_{l:j_{\text{top}}+1}^{(2)}\right)+\text{Params}\left(\mathcal{S}\right)+\text{Params}\left(f_{i_{\text{top}}:1}^{(1)} \right)$
    \If{$\text{Params}\left(i_{\text{top}}, j_{\text{top}}\right) \leq \epsilon$}
        \State Set $\left(i^*_{\text{top}}, j^*_{\text{top}}\right)=\left(i_{\text{top}}, j_{\text{top}}\right)$
        \State Set $\text{found} = \text{True}$
        \State $g(x)=f_{l:j^*_{\text{top}}+1}^{(2)} \circ \mathcal{S} \circ f_{i^*_{\text{top}}:1}^{(1)}$
    \Else
        \State Set $\mathcal{S}_{i_{\text{top}}j_{\text{top}}}=-\infty$
        \State Set $\text{found} = \text{False}$
    \EndIf
    \EndWhile
    \end{algorithmic}
\end{algorithm}

\textbf{How to stitch: the stitching directions.} In the previous subsection, we have obtained the optimal split point of two pre-trained models. Given two pre-trained models with different scales and complexities, there are two options for stitching them together: \textbf{Fast-to-Slow} and \textbf{Slow-to-Fast}. Taking \cref{fig:vis}(b) as an example, Fast-to-Slow takes the first portion of layers from a smaller and faster model and the second portion of layers from a larger and slower model. Slow-to-Fast goes in a reverse direction. For ease of understanding, we assume that $f^{(1)}\left(x ; \theta^{(1)}\right)$ is the larger, slower model, and $f^{(2)}\left(x ; \theta^{(2)}\right)$ is the smaller, faster model. Then the stitched model can be formulated as:
\begin{equation}
g(x) = \begin{cases}
f_{k:i^*_{\text{top}}+1}^{(1)} \circ \mathcal{S} \circ f_{j^*_{\text{top}}:1}^{(2)}, & \text{\quad Fast-to-Slow} \\
f_{l:j^*_{\text{top}}+1}^{(2)} \circ \mathcal{S} \circ f_{i^*_{\text{top}}:1}^{(1)}, & \text{\quad Slow-to-Fast} \\
\end{cases}
\end{equation}By default, ReStNet uses a Slow-to-Fast stitching direction, counter to conventional design principles that typically increase model width as depth grows. We provide detailed experiments in \cref{tab:stitchingType}.

\textbf{How to train the stitched model.} After obtaining the stitched model, it is usually necessary to fine-tune it to ensure that the model can adapt to the target task. As illustrated in \cref{fig:vis}(c), should we train the whole model or the partial part? However, directly fine-tuning the entire stitched model may lead to the following problems: 
\begin{itemize}
    \item Fine-tuning the entire model requires significant computing resources and time, especially for large models.
    \item Fine-tuning the entire model may overwrite the original pre-trained parameters, which have already learned general feature representations, resulting in a waste of computing resources.
\end{itemize}To address the above-mentioned problems, we introduce a partial-model fine-tuning strategy. Inspired by transfer learning~\cite{lee2022surgical,shen2021backdoor,zhu2021semantic}, where it is common to fine-tune only the parts close to the output layer (e.g., the last few layers or classifiers), we train only the stitched layer. This can be formulated as:
\begin{equation}
    g(x)=\overline{f_{l:j^*_{\text{top}}+1}^{(2)}} \circ \mathcal{S} \circ \overline{f_{i^*_{\text{top}}:1}^{(1)}},
\end{equation}where $\overline{f_{i^*_{\text{top}}:1}^{(1)}}$ and $\overline{f_{l:j^*_{\text{top}}+1}^{(2)}}$ denote the parameters of $f_{i^*_{\text{top}}:1}^{(1)}$ and $f_{l:j^*_{\text{top}}+1}^{(2)}$ are fixed and not updated. Finally, the stitched model updates its parameters using gradient descent by minimizing the loss $\ell$ (e.g., cross-entropy loss) on the target task $\mathcal{D}$: 
\begin{equation}
\mathcal{L}(\theta)=\mathbb{E}_{(X, Y) \sim \mathcal{D}}[\ell(g(X), Y)].
\end{equation}The detailed training algorithm is provided in \cref{algorithm:training} with PyTorch style. Although this paper focuses on training the stitching layer, we also provide experiments in \cref{sec:additional} on training the entire model. These experiments demonstrate the superiority of our partial-model fine-tuning strategy.

\begin{algorithm}[t]
    \caption{Model Training}
    \label{algorithm:training}
    \textbf{Input}: A stitched model $g(x)=f_{l:j^*_{\text{top}}+1}^{(2)} \circ \mathcal{S} \circ f_{i^*_{\text{top}}:1}^{(1)}$, dataset $\mathcal{D}$.\\
    \textbf{Output}: A Fine-tuned stitched model.\\
    \begin{algorithmic}[1] 
    \State Fixed the parameters of $f_{i^*_{\text{top}}:1}^{(1)} \Longrightarrow \overline{f_{i^*_{\text{top}}:1}^{(1)}}$
    \State Fixed the parameters of $f_{l:j^*_{\text{top}+1}}^{(1)} \Longrightarrow \overline{f_{l:j^*_{\text{top}+1}}^{(1)}}$
    \For{$i = 1$ to $n_{\text{iters}}$}
        \State Get next mini-batch of data $X$ and label $Y$.
        \State \textit{optimizer.zero\_grad()}. \Comment{Clear gradients}
        \State $\hat{Y}=g(x)=f_{l:j^*_{\text{top}}+1}^{(2)} \circ \mathcal{S} \circ \overline{f_{i^*_{\text{top}}:1}^{(1)}}(X)$. \Comment{Model Inference}
        \State \textit{loss=criterion($\hat{Y}$, $Y$)}. \Comment{Compute loss}
        \State \textit{loss.backward()}. \Comment{Compute gradients}
        \State \textit{optimizer.step()}. \Comment{Update weights}
    \EndFor
    \end{algorithmic}
\end{algorithm}

\section{Experiments}
\label{sec:Experiments}
\subsection{Experimental Setting}
\textbf{Datasets and Hyperparameters.} We conduct experiments on $5$ popular and publicly-available datasets, including CIFAR10~\cite{CIFAR}, CIFAR100~\cite{CIFAR}, DTD~\cite{DTD}, Oxford-IIIT Pets~\cite{Pets} and Imagenette~\cite{Howard_Imagenette_2019}. We first calculate the CKA similarity between two layers using a batch size of $256$. To ensure stability, we repeat the calculation $5$ times and take the average. Then we conduct model stitching. For training the CNN-CNN stitched models, we set the initial learning rate, batch size, weight decay, number of epochs, and momentum to $0.01$, $256$, $0.005$, $150$, and $0.9$, respectively. For training the Transformer-Transformer and CNN-Transformer stitched models, we set the initial learning rate, batch size, weight decay, number of epochs, and momentum to $0.00375$, $128$, $0.05$, $300$, and $0.9$, respectively.

\textbf{Pre-trained models.} For CNN-CNN, we fully train the base models (VGG11~\cite{VGG}, VGG19~\cite{VGG}, ResNet34~\cite{ResNet}) on the target dataset (CIFAR10, CIFAR100). Besides, for Transformer-Transformer and CNN-Transformer, we use the pre-trained base models (Swin-B~\cite{swin}, DeiT-S~\cite{deit}, DeiT-Ti~\cite{deit}, Swin-Ti~\cite{swin}, and ResNet18\cite{ResNet}) on ImageNet.

\textbf{Stitching layers.} After finding an appropriate position, we use a 1D convolutional layer and a linear layer as the stitching layer for CNN-CNN and Transformer-Transformer, respectively. As for CNN-Transformer, we first use a 2D convolutional layer and then flatten it to three dimensions.

\begin{table*}[htbp]
\caption{Performance and parameters trade-off of different homogeneous stitching ReStNet on various datasets.}
  \centering
    \begin{tabular}{c|c|cccccc}
    \toprule
    Stitching Type & Dataset & Front Model & Behind Model & Acc(\%) & Params & FLOPs & Trainable Params \\
    \midrule
    \multirow{20}[8]{*}{CNN-CNN} & \multirow{10}[4]{*}{CIFAR10} & -     & VGG11 & 92.26 & 9.76M & 153.90M & 9.76M \\
          &       & VGG19 & VGG11 & 93.31 & 10.53M & 267.54M & 0.59M \\
          &       & VGG19 & VGG11 & 93.70 & 11.12M & 305.39M & 2.36M \\
          &       & VGG19 & VGG11 & 93.91 & 20.56M & 399.86M & 2.36M \\
          &       & VGG19 & -     & 93.87 & 20.57M & 399.87M & 20.57M \\
\cmidrule{3-8}          &       & -     & VGG11 & 92.26 & 9.76M & 153.90M & 9.76M \\
          &       & ResNet34 & VGG11 & 92.52 & 9.98M & 438.46M & 0.07M \\
          &       & ResNet34 & VGG11 & 94.54 & 11.64M & 689.26M & 0.59M \\
          &       & ResNet34 & VGG11 & 95.09 & 12.23M & 1039.02M & 1.18M \\
          &       & ResNet34 & -     & 95.09 & 21.28M & 1163.51M & 21.28M \\
\cmidrule{2-8}          & \multirow{10}[4]{*}{CIFAR100} & -     & VGG11 & 67.25 & 9.80M  & 153.95M & 9.80M \\
          &       & VGG19 & VGG11 & 67.98 & 9.99M & 229.71M & 0.15M \\
          &       & VGG19 & VGG11 & 68.75 & 10.58M & 267.59M & 0.59M \\
          &       & VGG19 & VGG11 & 69.34 & 11.17M & 305.40M & 0.59M \\
          &       & VGG19 & -     & 70.39 & 20.61M & 399.92M & 20.61M \\
\cmidrule{3-8}          &       & -     & VGG11 & 67.25 & 9.80M  & 153.95M & 9.80M \\
          &       & ResNet34 & VGG11 & 68.58 & 10.92M & 636.49M & 0.15M \\
          &       & ResNet34 & VGG11 & 70.34 & 11.69M & 689.31M & 0.59M \\
          &       & ResNet34 & VGG11 & 71.45 & 12.87M & 764.94M & 0.59M \\
          &       & ResNet34 & -     & 76.68 & 21.33M & 1163.55M & 21.33M \\
    \midrule
    \multirow{10}[4]{*}{Transformer-Transformer} & \multirow{5}[2]{*}{DTD} & -     & DeiT-S & 70.48 & 21.68M & 4607.97M & 21.68M \\
          &       & Swin-B & DeiT-S & 72.02 & 35.92M & 9642.49M & 0.22M \\
          &       & Swin-B & DeiT-S & 76.17 & 44.01M & 11274.36M & 0.22M \\
          &       & Swin-B & DeiT-S & 77.71 & 68.26M & 16169.99M & 0.22M \\
          &       & Swin-B & -     & 77.13 & 86.79M & 15466.85M & 86.79M \\
\cmidrule{2-8}          & \multirow{5}[2]{*}{Pets} & -     & DeiT-Ti & 91.85 & 5.53M & 1258.23M & 5.53M \\
          &       & DeiT-S & DeiT-Ti & 92.45 & 16.80M & 3617.65M & 0.08M \\
          &       & DeiT-S & DeiT-Ti & 92.61 & 18.13M & 3924.94M & 0.08M \\
          &       & DeiT-S & DeiT-Ti & 93.92 & 19.02M & 4099.23M & 0.08M \\
          &       & DeiT-S & -     & 93.62 & 21.68M & 4607.97M & 21.68M \\
    \bottomrule
    \end{tabular}%

      \label{tab:main_experiment}%
\end{table*}%

\subsection{Main results}
\label{sec:main}
\textbf{Stitching CNNs (homogeneous stitching).} As shown in Table~\ref{tab:main_experiment}, we perform experiments on both CIFAR10 and CIFAR100 with combinations of VGG19, VGG11, and ResNet34. The results demonstrate that ReStNet consistently maintains competitive accuracy across different parameter budgets. On CIFAR10, VGG19-VGG11 progressively improves accuracy as the total parameters increase. For instance, using VGG19 as the front model and VGG11 as the behind model, the accuracy rises from $93.31\%$ to $93.98\%$ when the parameter count increases from $10.35$M to $12.90$M. A similar trend is observed when stitching ResNet34 with VGG11, where the accuracy improves from $94.01\%$ to $95.09\%$. These results demonstrate that adding more parameters via different layer combinations produces progressively better performance, and that ReStNet can generate a variety of hybrid models to trade off size and accuracy.

The same phenomenon is evident on CIFAR100. VGG19-VGG11 results in a steady improvement in accuracy from $67.98\%$ to $69.34\%$, indicating that ReStNet can achieve performance comparable to or even better than the larger model alone by effectively utilizing structural compatibility. When ResNet34 is used as the front model, accuracy grows as well as from $68.55\%$ to $71.45\%$, as the parameter budget increases. This again highlights ReStNet’s ability to produce diverse, resource-tailored networks that progressively refine performance.

\begin{figure}[t]
	\centering
    \centering
    \includegraphics[width=0.5\textwidth]{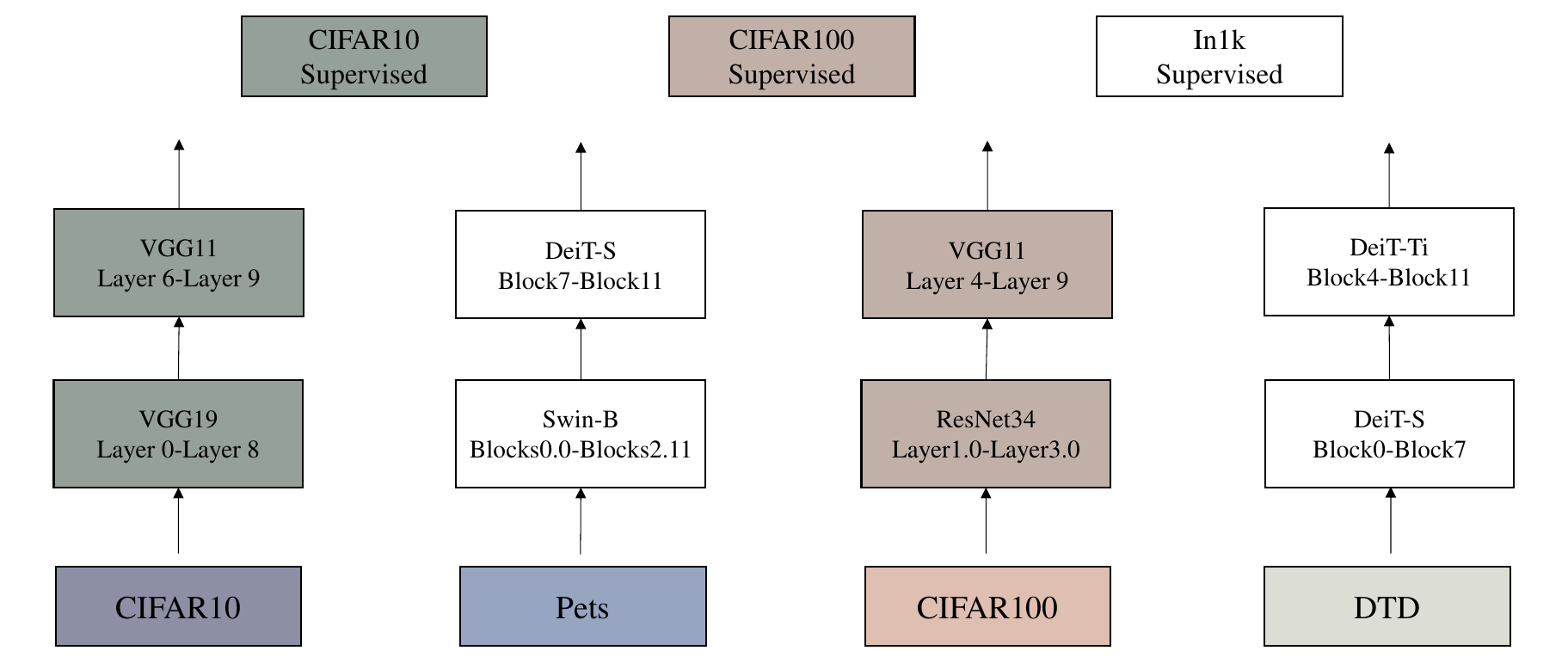} 
 \caption{Stitching models on different datasets.}
 \label{fig:stitch_structure}
\end{figure}

To better illustrate the stitching configuration, \cref{fig:stitch_structure} shows the layer-wise composition of selected stitched models. For example, on the CIFAR10 experiment, ReStNet stitches layers 0–8 from VGG19 with layers 6–9 from VGG11, selected automatically based on CKA similarity.

\textbf{Stitching Transformers (homogeneous stitching).} We next demonstrate ReStNet’s effectiveness on transformers by stitching pairs of Swin‐B, DeiT‐S, and DeiT‐Ti models on DTD and Pets. The results are shown in ~\cref{tab:main_experiment}. ReStNet maintains the same CKA‐guided method and reveals a clear, positive correlation between total parameters and accuracy.

On DTD, the single DeiT‐S achieves $70.48\%$ accuracy. By growing parameter constraints, Swin-B-DeiT-S with $35.92$M, $44.01$M, and $68.26$M parameters reach $72.02\%$, $76.17\%$, and $77.71\%$ respectively. Notably, when the parameters approach that of Swin‐B, the Swin-B-DeiT-S even surpasses the original Swin‐B performance, demonstrating that ReStNet is able to utilize the complementary strengths of both models. 

A similar trend emerges on Pets: the base DeiT‐Ti starts at $91.85\%$ accuracy, and DeiT-S-DeiT-Ti of $16.80$M, $18.13$M, and $21.68$M parameters improve to $92.45\%$, $92.61\%$, and $93.62\%$. Again, ReStNet not only bridges the gap between lightweight and heavyweight models but can exceed the performance of the high-capacity parent, illustrating its capacity to generate diverse, high-performing hybrid models.

Beyond the accuracy, we also visualize the ReStNet configuration to understand which transformer blocks are selected. \cref{fig:stitch_structure} presents two representative stitching schemes. For instance, in the Pets, ReStNet pairs Swin-B’s blocks 0.0–2.11 with DeiT-S’s blocks 7–11. This visualization makes clear the internal configuration of the ReStNet.

\textbf{Stitching CNN and Transformer (heterogeneous stitching).} To further validate the generalizability of ReStNet, we experiment with a heterogeneous CNN-Transformer by stitching ResNet18 and Swin‐Ti on the ImageNette dataset.

We select ResNet18 as the CNN front model and Swin-S as the Transformer back model to form ResNet18-Swin-Ti. Following the same method, we first calculate the layer-wise CKA similarity between the two models. And then, under a given parameter constraint $\epsilon=26.74M$, we identify the highest-similar layer as the stitching position.

\cref{tab:CNN_transformer} demonstrates that the ResNet18-Swin-Ti even surpasses the performance of Swin-S remarkably. The ResNet18-Swin-Ti achieves an accuracy improvement of $0.23\%$ over Swin-Ti, while having fewer total parameters. Meanwhile, there is no doubt that the ResNet18-Swin-Ti has only trainable parameters compared with ResNet18 and Swin-Ti.

Despite their architectural differences, the ResNet18-Swin-Ti still delivers competitive performance. This finding demonstrates that ReStNet can effectively bridge structurally distinct models to produce a diverse set of high-performing hybrid models. Moreover, it confirms the feasibility of cross-architecture stitching and underscores the broad applicability of our CKA-guided strategy beyond homogeneous scenarios.
\begin{table}[t]
  \caption{Performance and parameters trade-off of different heterogeneous stitching ReStNet.}
  \centering
\begin{tabular}{c|ccc}

    \toprule
    Dataset & Model & Acc(\%) & Trainable / Total Params \\
    \midrule
    \multirow{3}[2]{*}{Imagenette} & ResNet18 & 97.78 & 11.18M/11.18M \\
          & ResNet18-Swin-Ti & 99.72 & 0.03M/26.74M \\
          & Swin-Ti & 99.49 & 27.53M/27.53M \\
    \bottomrule
    \end{tabular}%

    \label{tab:CNN_transformer}%
\end{table}%

\subsection{Additional Experiments}
\label{sec:additional}
\textbf{Fast-to-Slow or Slow-to-Fast.} In \cref{sec:Method}, We propose to stitch the back half of the small model with the front half of the large model. To further investigate the impact of stitching order,  we conduct experiments on the CIFAR10 using ResNet34 (as the larger model) and VGG11 (as the smaller model). We evaluate two stitching directions: Slow-to-Fast (larger model followed by smaller model) and Fast-to-Slow (small model followed by large model). As shown in \cref{tab:stitchingType}, Slow-to-Fast consistently achieves higher accuracy with fewer total parameters compared to Fast-to-Slow. For instance,  a Slow-to-Fast with $12.23$M parameters achieves $95.09\%$ accuracy, outperforming a Fast-to-Slow with $19.98$M parameters and only $92.00\%$ accuracy. This result suggests that the Slow-to-Fast configuration has more potential.

\begin{table}[t]
  \caption{Comparison of the accuracy and number of parameters of Fast-to-Slow and Slow-to-Fast stitching directions on CIFAR10.}
  \centering
  \resizebox{\linewidth}{!}{
    \begin{tabular}{c|c|ccc}
    \toprule
    Dataset & Type  & Acc(\%)   & Trainable Params  & Params  \\
    \midrule
    \multirow{8}[4]{*}{CIFAR10} & \multirow{4}[2]{*}{Fast-to-Slow} & 83.22 & 0.07M  & 21.21M \\
          &       & 90.45 & 0.59M  & 20.58M \\
          &       & 92.00    & 1.18M  & 19.98M \\
          &       & 92.32 & 1.18M  & 21.16M \\
\cmidrule{2-5}          & \multirow{4}[2]{*}{Slow-to-Fast} & 92.52 & 0.07M  & 9.98M \\
          &       & 94.54 & 0.59M  & 11.64M \\
          &       & 95.09 & 1.18M  & 12.23M \\
          &       & 95.14 & 1.18M  & 13.41M \\
    \bottomrule
    \end{tabular}%
  }

    \label{tab:stitchingType}
\end{table}%

\textbf{Full-model fine-tuning or Partial-model fine-tuning.} In \cref{sec:Method}, we choose to fine-tune the stitching layer. However, we truly have four fine-tuning strategies: 1) Only fine-tuning the stitching layer. 2) Fine-tuning the stitching layer and the latter half of the model. 3) Fine-tuning the stitching layer and the front half of the model. 4) Fine-tuning the whole model. To evaluate the trade-off between performance and training cost, we choose VGG19-VGG11 as the representative stitched model and perform all experiments on CIFAR10. The results are summarized in \cref{tab:training_strategy}. As it is shown, all strategies achieve comparable performance in accuracy. However, the stitching-only approach greatly reduces the number of trainable parameters compared with others. Specifically, in the second row of \cref{tab:training_strategy}, the approach, only training the stitching layer, reaches $93.31\%$, better than training the stitching layer and the latter half of the model, and training the stitching layer and the front half of the model. The stitching-only approach achieves near-optimal performance with minimal trainable parameters. This shows that the stitching-only approach can offer an ideal trade-off between performance and deployment cost.

\begin{table*}[htbp]
  \caption{Comparison of four fine‐tuning strategies for the VGG19–VGG11 on CIFAR10.}
  \centering
    \begin{tabular}{cc|cc|cc|cc|c}
    \toprule
    \multicolumn{8}{c}{Training Type}                             & \multirow{3}[6]{*}{Params} \\
\cmidrule{1-8}    \multicolumn{2}{c|}{Stitching Layer Only} & \multicolumn{2}{c|}{Stitching Layer + Latter} & \multicolumn{2}{c|}{Stitching Layer + Front} & \multicolumn{2}{c}{Whole Model} &  \\
\cmidrule{1-8}    Acc(\%)   & Trainable Params (M) & Acc(\%)   & Trainable Params & Acc(\%)   & Trainable Params  & Acc(\%)   & \multicolumn{1}{c}{Trainable Params } &  \\
    \midrule
    92.65 & 0.15M & 93.22 & 9.83M & 92.32 & 0.26M & 93.46 & 9.94M & 9.94M \\
    93.31 & 0.59M & 92.77 & 9.97M & 92.65 & 1.15M & 93.24 & 10.53M & 10.53M \\
    93.70  & 2.36M & 93.69 & 7.61M & 94.00    & 5.87M & 94.18 & 11.12M & 11.12M \\
    93.83 & 0.26M & 93.93 & 0.53M & 93.91 & 20.56M & 93.96 & 20.83M & 20.83M \\
    93.91 & 2.36M & 93.92 & 2.89M & 94.15 & 20.03M & 93.98 & 20.56M & 20.56M \\
    \bottomrule
    \end{tabular}%

    \label{tab:training_strategy}%
\end{table*}%

\begin{figure*}[htbp]
	\centering 
    \centering
    \includegraphics[width=0.99\textwidth]{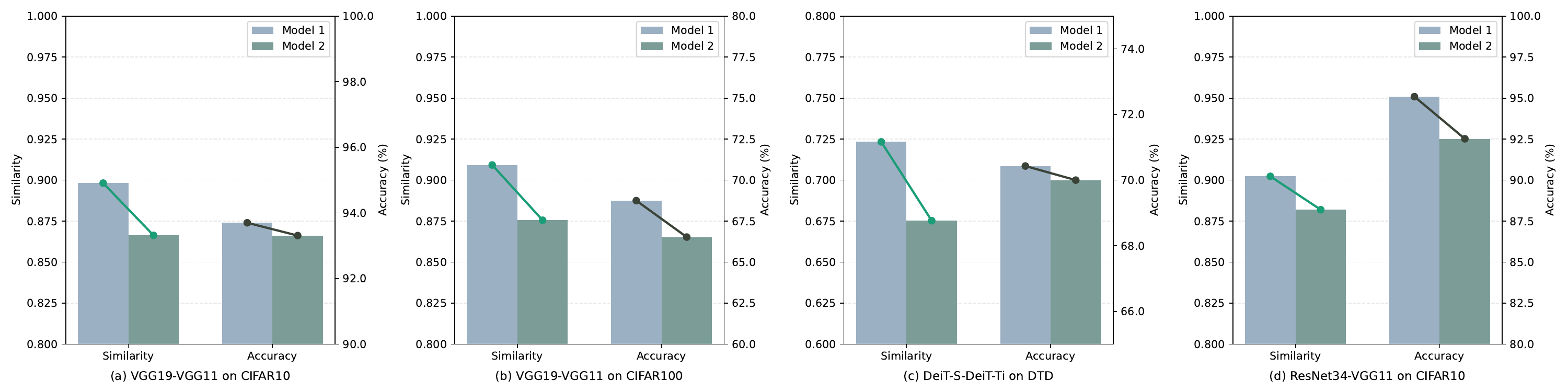} 
 \caption{The CKA similarity and corresponding accuracy of the stitched model for different candidate stitching points.}
 \label{fig:similarity_bar_with_lines}
\end{figure*}

\textbf{Why should we stitch the most similar layers.} In model stitching, how to select the optimal layers for stitching is also a central question. In \cref{sec:Method}, under resource constraints,  we choose the position with maximum CKA similarity value. In this section, we select a variety of models, including CNNs (VGG11, VGG19, ResNet34) and transformers (DeiT-S, and DeiT-Ti). Besides, we conduct experiments on multiple datasets such as CIFAR10, CIFAR100, and DTD. In the experiments, we only change the similarity between the layers, keeping other settings consistent. \cref{fig:similarity_bar_with_lines} summarizes the results. Across all datasets and model pairs, we observe a consistent positive correlation between CKA similarity and final task accuracy. Specifically, stitching layers with the higher similarity often yields the better performance. These results strongly demonstrate that CKA similarity can be effectively used as an indicator for model stitching position selection.

\begin{figure*}[t]
	\centering
    \centering
    \includegraphics[width=0.99\textwidth]{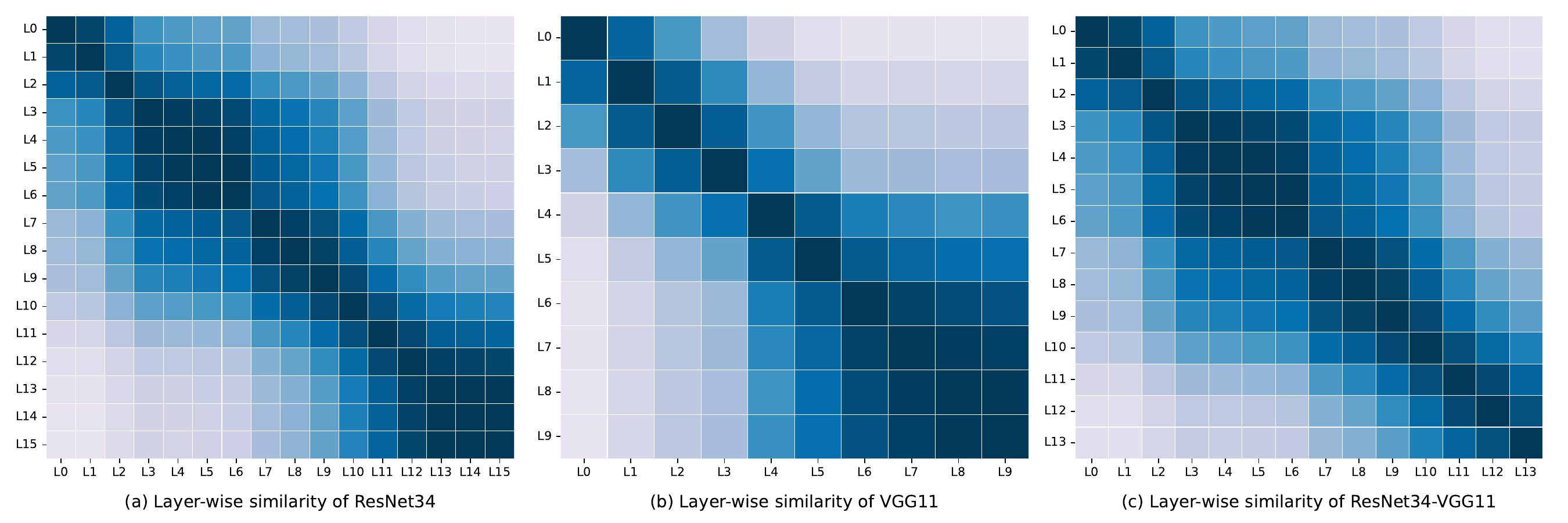} 
 \caption{Visualization of the layer-wise CKA similarity for the base pre-trained models and ReStNet. (a) and (b) show the layer-wise similarity of base models---ResNet34 and VGG11, respectively. (c) shows the layer-wise similarity of the ReStNet.}
 \label{fig:heatmap}
\end{figure*}

\textbf{Visualize CKA similarity} To further investigate the internal representation of the stitching model, we visualize the CKA similarity matrices on different configurations. Specifically, we first visualize the CKA similarity matrices of two pre-trained models---ResNet34 and VGG11. And then visualize the matrix of ResNet34-VGG11.


As shown in \cref{fig:heatmap}, ResNet34 and VGG11 each exhibit distinctive CKA patterns: ResNet34’s highest similarities lie along its diagonal, reflecting consistent feature progression across all depths, while VGG11 shows stronger correlations concentrated in its deeper layers. In contrast, the ReStNet’s displays a markedly different pattern. 
Instead of a single straight diagonal, the ReStNet's heatmap splits into two diagonal segments: one following ResNet34’s early-layer pattern, and a second following VGG11’s deep-layer pattern, clearly marking the stitch point. Furthermore, we observe new bands of moderate similarity around ResNet34’s block boundaries and VGG11’s, reflecting hybrid representations that neither parent network exhibited alone. These detailed shifts confirm that the ReStNet forms a new internal representation, highlighting the need for careful layer selection and fine-tuning.

\section{Conclusion}
\label{sec:Conclusion}
In this paper, we present ReStNet, a reusable and stitchable network that enables efficient adaptation of pre-trained models to dynamic resource constraints. ReStNet identifies the most optimal stitching positions between two models by computing CKA similarity and constructs a hybrid network by retaining the early layers of the model with greater capacity and appending the deep layers of the smaller model. ReStNet can achieve competitive performance with negligible cost by only fine-tuning the stitching layer. Finally, ReStNet supports both homogeneous and heterogeneous model stitching and achieves competitive performance on various datasets, demonstrating its potential for flexible deployment in dynamically changing resource-constrained IoT scenarios.

%
\bibliographystyle{IEEEtran}
\bibliography{reference}

\begin{IEEEbiography}[{\includegraphics[width=1in,height=1.25in,clip,keepaspectratio]{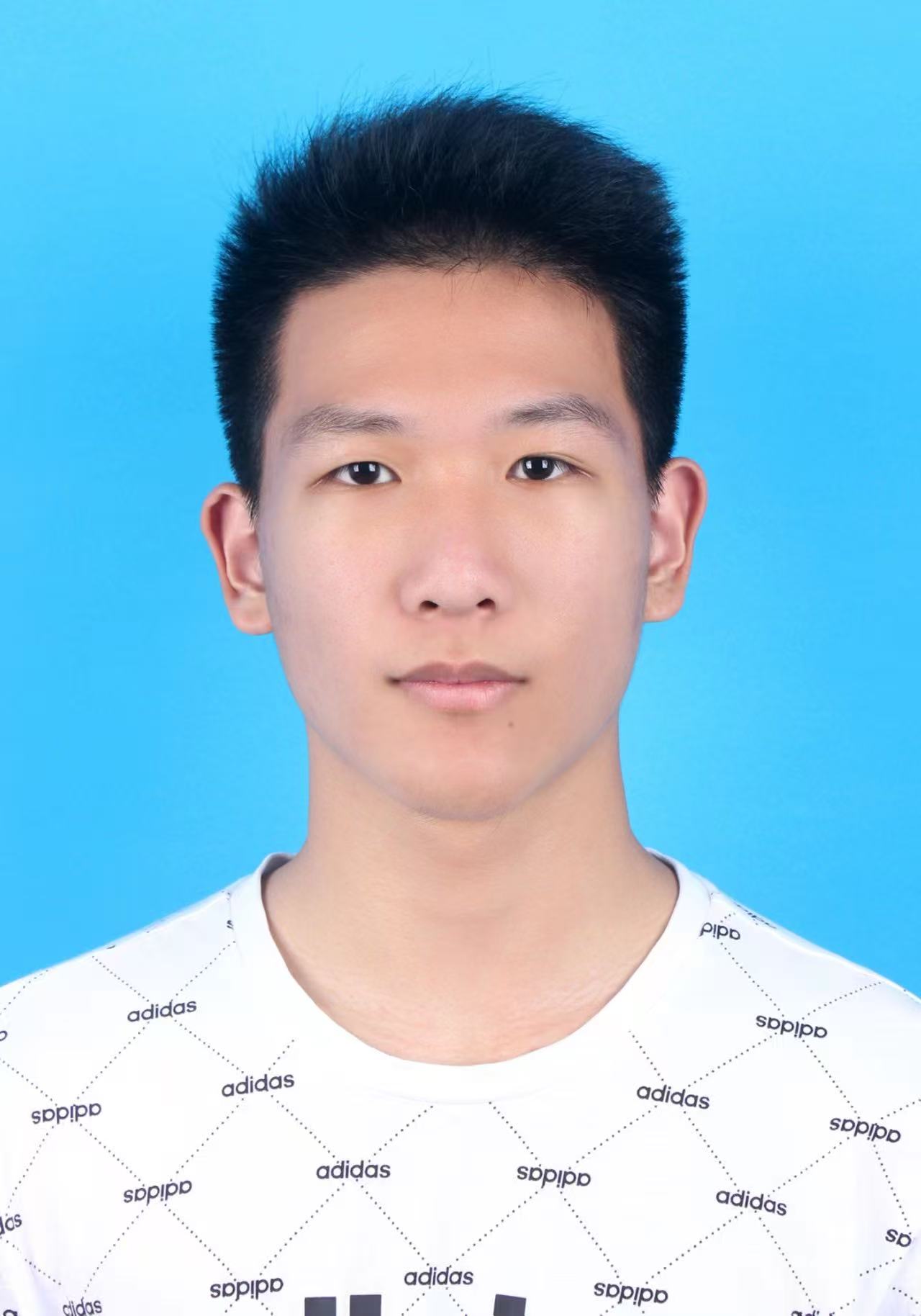}}]{Maoyu Wang}
is currently pursuing the B.S. degree at the College of Computer Science and Technology, Zhejiang University of Technology, Hangzhou, China. His research interests include deep learning and computer vision.
\end{IEEEbiography} 

\begin{IEEEbiography}[{\includegraphics[width=1in,height=1.25in,clip,keepaspectratio]{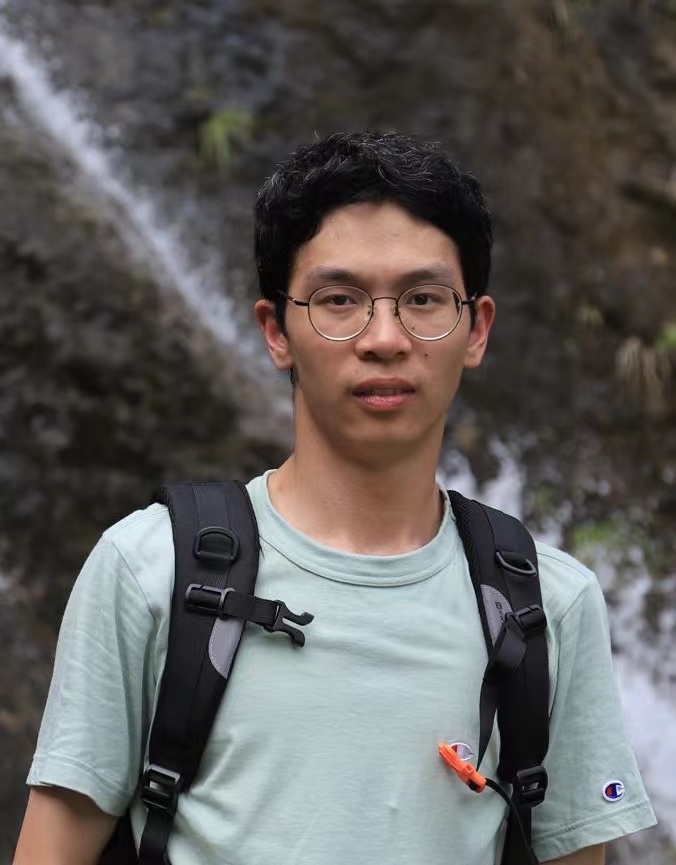}}]{Yao Lu}
received his B.S. degree from Zhejiang University of Technology and is currently pursuing a Ph.D. in control science and engineering at Zhejiang University of Technology. He is a visiting scholar with the Centre for Frontier AI Research, Agency for Science, Technology and Research, Singapore. He has published several academic papers in international conferences and journals, including ECCV, TNNLS, Neurocomputing, Information Sciences and TCCN. He serves as a reviewer of ICLR2025, CVPR2025, ICCV25, TNNLS and JSAC. His research interests include deep learning and computer vision, with a focus on artificial intelligence and model compression.
\end{IEEEbiography} 

\begin{IEEEbiography}[{\includegraphics[width=1in,height=1.25in,clip,keepaspectratio]{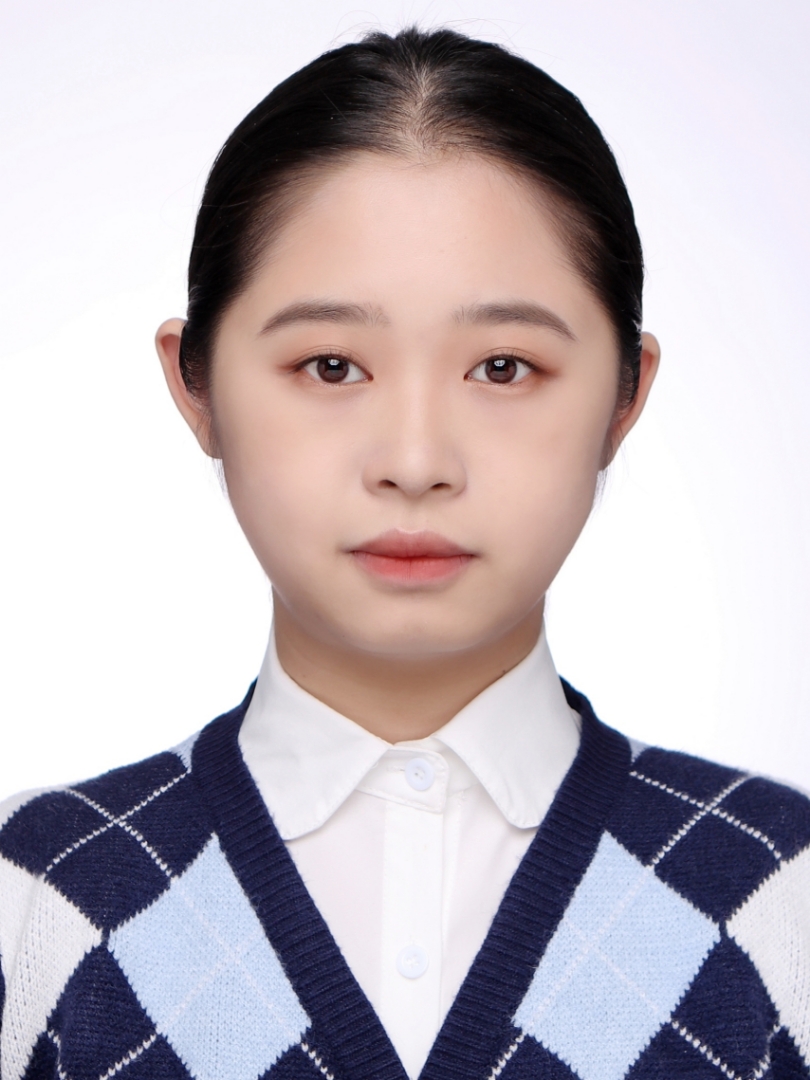}}]{Jiaqi Nie} is currently pursuing the Eng.D degree in electronic information with the Institute of Cyberspace Security, Zhejiang University of Technology, Hangzhou, China. Her research interests include complex networks and data analysis.
\end{IEEEbiography}

\begin{IEEEbiography}[{\includegraphics[width=1in,height=1.25in,clip,keepaspectratio]{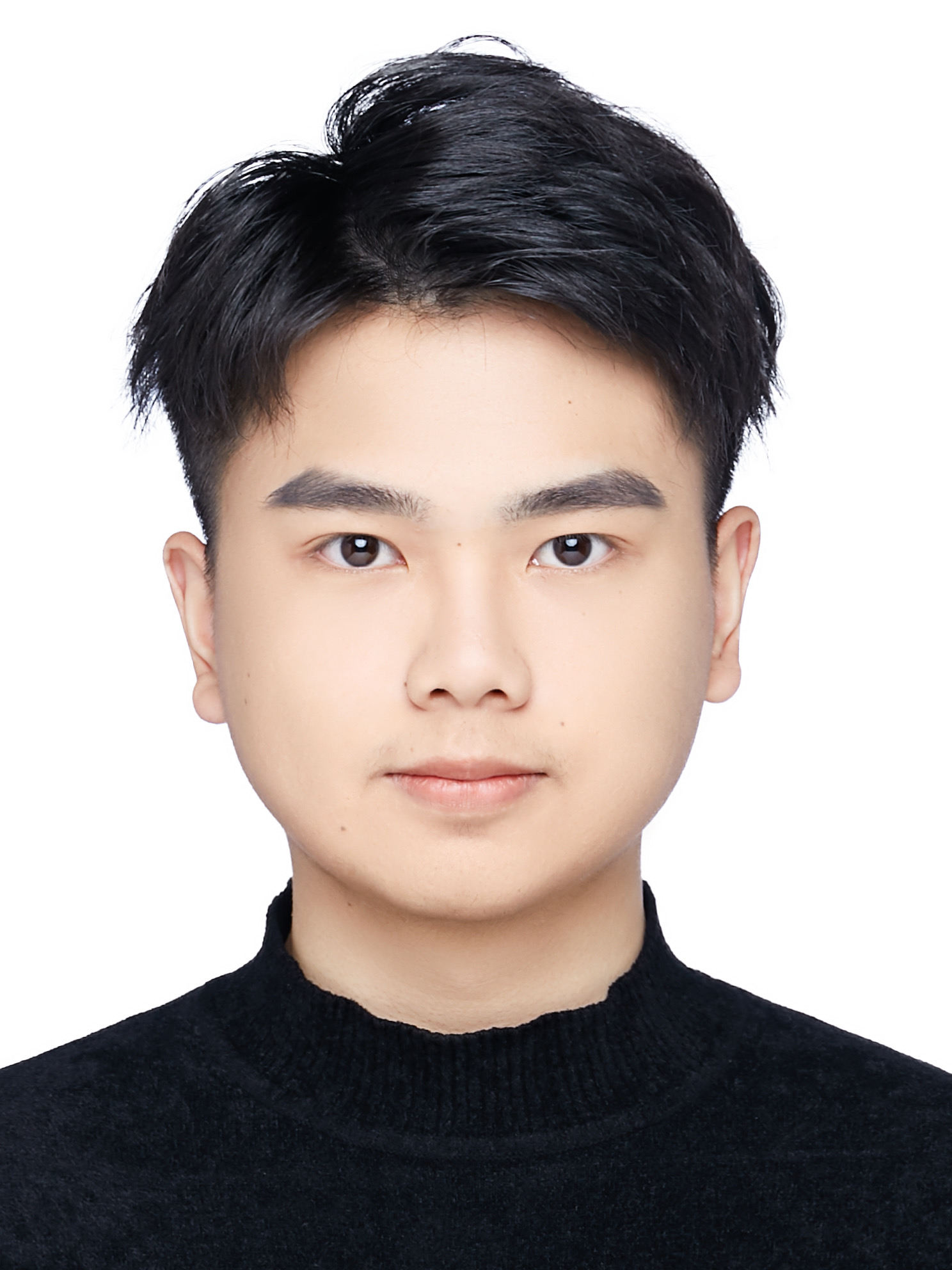}}]{Zeyu Wang} received the BS degree from Shaoxing University, Shaoxing, China, in 2021. He is currently pursuing the Ph.D. degree at the College of Information Engineering, Zhejiang University of Technology, Hangzhou, China. His current research interests include graph data mining, recommender system and bioinformatics.
\end{IEEEbiography}

\begin{IEEEbiography}[{\includegraphics[width=1in,height=1.25in,clip,keepaspectratio]{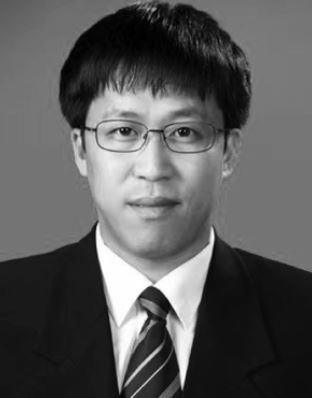}}]{Yun Lin}
(Member, IEEE) received the B.S. degree in electrical engineering from Dalian Maritime University, Dalian, China, in 2003, the M.S. degree in communication and information system from the Harbin Institute of Technology, Harbin, China, in 2005, and the Ph.D. degree in communication and information system from Harbin Engineering University, Harbin, in 2010. From 2014 to 2015, he was a Research Scholar with Wright State University, Dayton, OH, USA. He is currently a Full Professor with the College of Information and Communication Engineering, Harbin Engineering University. He has authored or coauthored more than 200 international peer-reviewed journal/conference papers, such as IEEE Transactions on Industrial Informatics, IEEE Transactions on Communications, IEEE Internet of Things Journal, IEEE Transactions on Vehicular Technology, IEEE Transactions on Cognitive Communications and Networking, TR, INFOCOM, GLOBECOM, ICC, VTC, and ICNC. His current research interests include machine learning and data analytics over wireless networks, signal processing and analysis, cognitive radio and software-defined radio, artificial intelligence, and pattern recognition.
\end{IEEEbiography}

\begin{IEEEbiography}[{\includegraphics[width=1in,height=1.25in,clip,keepaspectratio]{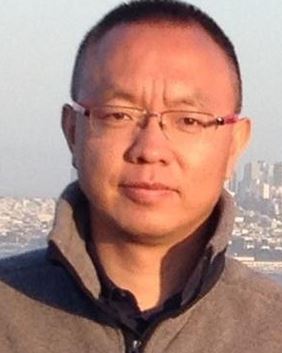}}]{Qi Xuan}
(Senior Member, IEEE) received the B.S. and Ph.D. degrees in control theory and engineering from Zhejiang University, Hangzhou, China, in 2003 and 2008, respectively. He was a Postdoctoral Researcher with the Department of Information Science and Electronic Engineering, Zhejiang University from 2008 to 2010, and a Research Assistant with the Department of Electronic Engineering, City University of Hong Kong, Hong Kong, in 2010 and 2017, respectively. From 2012 to 2014, he was a Postdoctoral Fellow with the Department of Computer Science, University of California at Davis, Davis, CA, USA. He is currently a Professor with the Institute of Cyberspace Security, College of Information Engineering, Zhejiang University of Technology, Hangzhou, and also with the PCL Research Center of Networks and Communications, Peng Cheng Laboratory, Shenzhen, China. He is also with Utron Technology Company Ltd., Xi’an, China, as a Hangzhou Qianjiang Distinguished Expert. His current research interests include network science, graph data mining, cyberspace security, machine learning, and computer vision.
\end{IEEEbiography}

\begin{IEEEbiography}[{\includegraphics[width=1in,height=1.25in,clip,keepaspectratio]{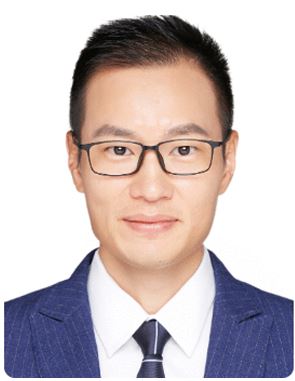}}]{Guan Gui}
    received the Dr. Eng. degree in information and communication engineering from the University of Electronic Science and Technology of China, Chengdu, China, in 2012. From 2009 to 2012, with financial support from the China Scholarship Council and the Global Center of Education, Tohoku University, he joined the Wireless Signal Processing and Network Laboratory (Prof. Adachis laboratory), Department of Communications Engineering, Graduate School of Engineering, Tohoku University, as a Research Assistant and a Post Doctoral Research Fellow, respectively. From 2012 to 2014, he was supported by the Japan Society for the Promotion of Science Fellowship as a Post Doctoral Research Fellow with the Wireless Signal Processing and Network Laboratory. From 2014 to 2015, he was an Assistant Professor with the Department of Electronics and Information System, Akita Prefectural University. Since 2015, he has been a Professor with the Nanjing University of Posts and Telecommunications, Nanjing, China. He is currently involved in the research of big data analysis, multidimensional system control, super-resolution radar imaging, adaptive filter, compressive sensing, sparse dictionary designing, channel estimation, and advanced wireless techniques. He received the IEEE International Conference on Communications Best Paper Award in 2014 and 2017 and the IEEE Vehicular Technology Conference (VTC-spring) Best Student Paper Award in 2014. He was also selected as a Jiangsu Special Appointed Professor, as a Jiangsu High-Level Innovation and Entrepreneurial Talent, and for 1311 Talent Plan in 2016. He has been an Associate Editor of the Wiley Journal Security and Communication Networks since 2012.
\end{IEEEbiography}


\vfill

\end{document}